\pdfoutput=1

\documentclass[11pt]{article}

\usepackage[]{ACL2023}

\usepackage{times}
\usepackage{latexsym}

\usepackage{amsmath}  
\usepackage{amssymb}  

\usepackage{graphicx}
\usepackage{algorithm}
\usepackage{algpseudocode}
\usepackage{amsmath}
\DeclareMathOperator*{\argmax}{arg\,max}

\usepackage[T1]{fontenc}

\usepackage[utf8]{inputenc}

\usepackage{microtype}

\usepackage{inconsolata}

\title{Enhancing Long-Term Memory using Hierarchical Aggregate Tree for Retrieval Augmented Generation}

\author{Aadharsh Aadhithya A$^*$, Sachin Kumar S$^*$ \and Soman K.P$^*$\\
 Amrita School of Artificial Intelligence,Coimbatore \\ 
 Amrita Vishwa Vidyapeetham, India  \\ }

\begin{document}
\maketitle
\footnotetext{Drafted: December 18,2023\\
    Arxiv Submission: \today}
\begin{abstract}

Large language models (LLMs) have limited context capacity, hindering reasoning over long conversations. We propose the Hierarchical Aggregate Tree (HAT) memory structure to recursively aggregate relevant dialogue context through conditional tree traversals. HAT encapsulates information from children nodes, enabling broad coverage with depth control. We formulate finding best context as optimal tree traversal. Experiments show HAT improves dialog coherence and summary quality over baseline contexts, demonstrating the technique's effectiveness for multi-turn reasoning without exponential parameter growth. HAT balances information breadth and depth for long-form dialogues. This memory augmentation enables more consistent, grounded long-form conversations from LLMs.
\end{abstract}

\section{Introduction}
Large language models (LLMs) like ChatGPT are having an impact across various areas and applications. One of the most straightforward applications is using LLMs as personalized chat agents. There have been several efforts to develop chatbots for various applications, both generic and domain-specific, particularly after the advent of LLMs and associated Pythonic libraries, which have made it very easy for people to develop their own chatbots.\\

\noindent Customizing and aligning these LLMs is still an active area of research. One basic alignment we would want is to make chatbots behave according to our expectations, particularly when context is specific and requires some information to be highlighted that is not necessarily in the model's pretraining corpus. While LLMs are considered snapshots of the internet, one limitation is that they are closed systems and providing external information to LLMs is an active research area.\\

Two primary ways of providing external data to LLMs are through: a) finetuning and b) retrieval-augmented generation (RAG). Finetuning requires access to model weights and can only be done with open models. RAG relies on strategies to retrieve information from a datastore given a user query, without needing internal model information, allowing it to be used with more model types. However, RAG is limited by the model's context length budget. Although very large context LLMs exist, the budget remains limited. Hence, how and what data is retrieved given a user query is an important research task.With the advent of "LLM agents", a separate memory management module is often required. Some solutions train models for this task while others take a retrieval-based approach. Still others use the LLM itself to accomplish it. However, current approaches tend to rely solely on providing a summary versus retrieving from a datastore, with little in between. We hence propose a method that combines both worlds using a new data structure called the "Hierarchical Aggregate Tree".

\subsection{Recent works}

There has been growing interest in developing techniques to enhance LLMs' capabilities for long-term multi-session dialogues. \cite{xu_beyond_2021} collected a dataset of human conversations across multiple chat sessions and showed that existing models perform poorly in maintaining consistency, necessitating long-context models with summarization and retrieval abilities. \cite{xu_long_2022} proposed a model to extract and update conversational personas over long-term dialogues. \cite{bae_keep_2022} presented a task and dataset for memory management in long chats, dealing with outdated information. \cite{wang_recursively_2023} proposed recursive summarization for equipping LLMs with long-term memory. \cite{lee_prompted_2023} showed prompted LLMs can match fine-tuned models for consistent open-domain chats. \cite{zhang_history-aware_2023} encoded conversation history hierarchically to generate informed responses. Developing more sophisticated strategies for information integration remains an open challenge.

\section{Task Definition}

The Task at hand is straightforward. Given a history of conversations between a 
user and an assistant(system), we are to predict the response of the system. In other words, given the history of conversation at time step t, $H_t = \{ u_1,a_1, u_2,a_2 \cdots a_{t-1}\}$(where $u_i$ represents a user response and $a_i$ represents assistant response) and a user query $u_t$, our task is to find a relevant function $f$ such that

$$a_t \approx LLM(u_t , f(H_t|u_t))$$

Where $f$ can be thought of as some function, mapping the entire history of conversation to a \textit{condensed space} conditioned on the user query at time step t. Note that, $f$ can be a selection function, a trained neural network as a memory agent, or even a simple summary function. In our experiments, the dataset is organized as sessions and episodes. An episode consists of multiple consecutive dialogue sessions with a specific user, oftentimes requiring information from previous sessions to respond back. Hence in addition to the history $H$, at the end of every session, we also have a memory $M^e_s$ for session $s$ and episode $e$, constructed by combining $H^e_s$ and $M^{e}_{s-1}$, where $H_s^e$ represents the history of session s and episode e. Therefore, we also have this auxiliary task of finding $M^e_s$ given $M^e_{s-1}$ and $H_s^e$ . 
\begin{figure*}[t]
    \centering
    \includegraphics[width=0.8\textwidth]{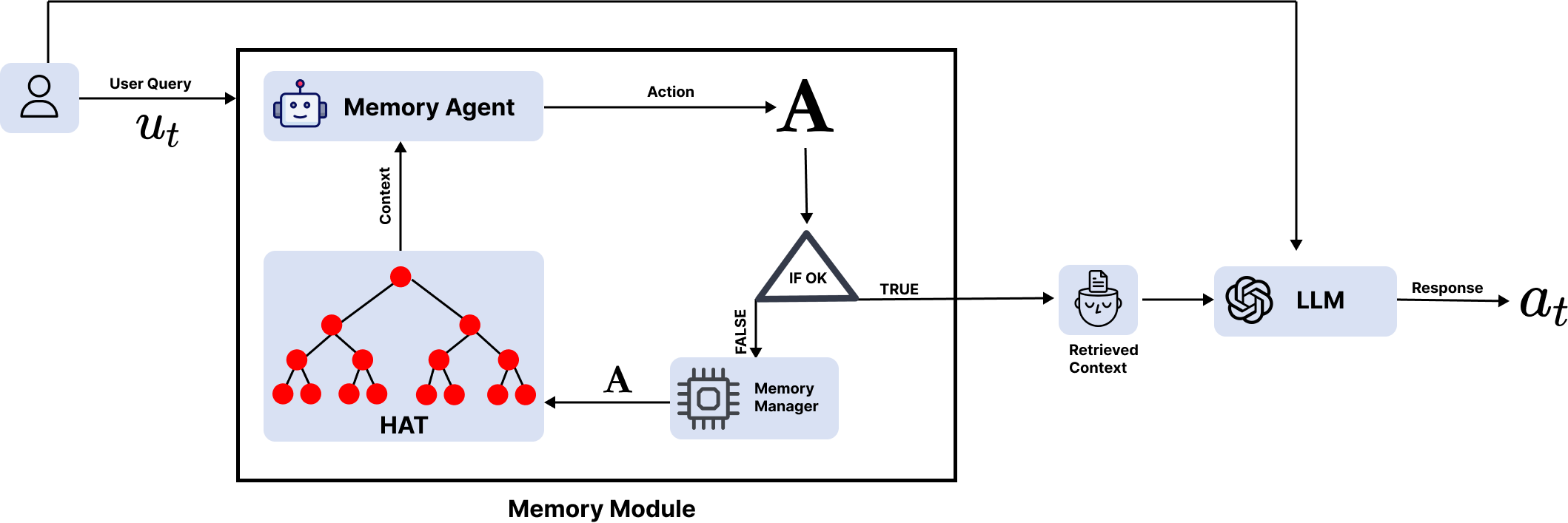}
    \caption{Overview of our approach. Given a user query, the memory module is responsible to give a relevant context by traversing the HAT. The LLM then generates response for the user query, given the context. }
    \label{fig:methedology}
\end{figure*} 

\section{Methedology}

\begin{figure}[t]
    \centering
    \includegraphics[width=0.4\textwidth]{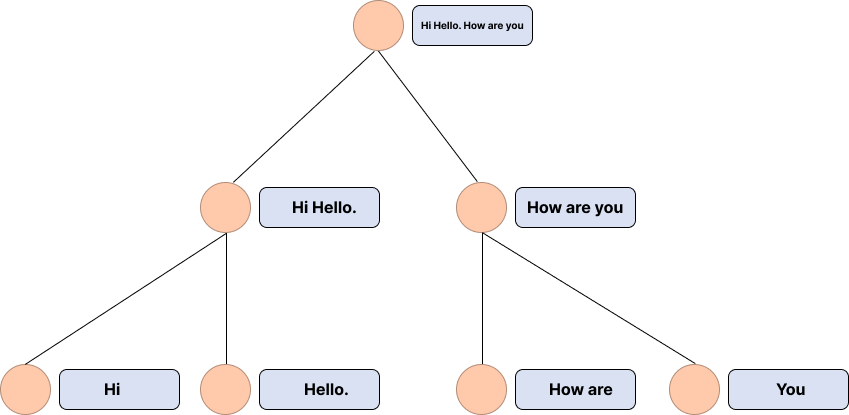}
    \caption{Illustration of HAT, with example aggregation function as simple concatenation and memory length of 2. }
    \label{fig:HAT}
\end{figure}

An overview of our methodology is depicted in Figure \ref{fig:methedology}. The memory module is tasked to retrieve a relevant context that has necessary information to respond to user query. The module does so by leveraging 2 primary components:
\begin{enumerate}
    \item \textbf{Hierarchical Aggregate Tree (HAT) :}  We propose the use of a novel datastructure called HAT to manage memory, particularly in long form texts like open domain conversations. It has some important property like \textit{resolution retention} as detailed in section \ref{sec:HAT}. At a high-level, the intended characteristic of HAT is that we have more resolution as we move Top-Down and we have more latest information as we move Left-Right.
    \item \textbf{Memory Agent :}  The memory agent is tasked to find the best traversal in HAT, conditioned on the user query such that at the end of traversal, we are in a node that contains relevant context to answer the user query.
\end{enumerate}

\subsection{Hierarchical Aggregate Tree (HAT)}
\label{sec:HAT}

Hierarchical Aggregate Tree (HAT) is defined as $HST = (L, M, A, \Sigma)$ , where, $L = \{l_0, l_1, \ldots, l_n\}$ is a finite set of layers,$M$ is the \textit{memory length}, a positive integer,$A$ is an aggregation function and $\Sigma$ is a set of nodes. The layers in L are hierarchical, with $l_0$ being the root layer. Each layer $l_i \in L$ is a set of nodes $\Sigma_i$, where $\Sigma_i \subseteq \Sigma$. A node $\sigma \in \Sigma$ can have children nodes and contains a text element. A node recalculates its text property, whenever there is a change in the node's children. The text property of a node $\sigma \in \Sigma_i \, i \neq |L|$ is given by $A(C(\sigma))$ , where $C(\sigma) = \{  \tau \in \Sigma \mid \tau $ is a child of $ \sigma \}$ \\ 



\subsubsection*{Aggregate Function}
The aggregate function $A$ maps the text of child nodes to produce the text stored at a parent node. $$A: \mathcal{P}(\Sigma) \rightarrow Text$$

Where, $\mathcal{P}(\Sigma)$ represents the power set of nodes. The exact implementation of $A$ can vary depending on the usecase. Figure \ref{fig:HAT} for example, depicts an HAT with concatenation as aggregate function. For our implementation, we use GPT as aggregate function.

It is important that the aggregate function should be designed to produce concise summaries reflecting key information from child nodes for a meaningful HAT. It executes whenever new child nodes are inserted to maintain consistency up the tree. For our implementation, we use GPT as aggregate function. We say GPT to summarize persona's from the children's text. The exact prompt given can be found in the appendix.
\subsubsection*{Node}
A node $\sigma \in \Sigma$ represents a single element in the HAT structure. Whenever the set of children nodes for $\sigma$ changes, the \texttt{update\_text()} method is called to update $\sigma$'s text. This text is given by applying the aggregator function to the set of texts from the current child nodes.  The previous aggregated texts of given different combinations of children are cached in \texttt{previous\_complete\_state} to enable reuse instead of recomputing. \\

After updating, $\sigma$ triggers the parent node to also update, thereby propagating changes upwards in the HAT structure. \\

Each node $\sigma$ contains the following components:
\begin{itemize}
\item \texttt{id}: A unique identifier
\item \texttt{text}: The node's aggregated text content
\item \texttt{previous\_complete\_state}: A dictionary mapping hashes of the node's children to the previously aggregated text when the node had those children
\item \texttt{parent}: The parent node in the HAT (None for the root node)
\item \texttt{aggregator}: The aggregation function $A$
\item \texttt{children}: The set of child nodes $C(\sigma)$
\end{itemize}
\noindent  The HAT datastructure satisfies the invariant that if
$\sigma_{k,i} \in \Sigma_k$ is a child of $\tau_{k-1,j} \in \Sigma_{k-1} \leftrightarrow j = \lfloor \frac{i}{M} \rfloor$, where $\sigma_{k,i}$ is ith node in kth layer. This connects child nodes to parent nodes between layers based on the memory length, $M$. When inserting a new node $\phi \in \Sigma_y$, the aggregation function $A$ is recursively applied to update ancestor nodes. That is, For all $\sigma_{y-1,z} \in P(\phi)$ that are parents of $\phi$, $\sigma_{y,z}.text = A(C(\sigma_{y,z}))$,
Where $C(\sigma_{y,z}) = \{\tau \in \Sigma | \tau$ is a child of $\sigma_{y,z}\}$

This maintains the invariant while propagating updated information through the tree.
The number of layers $|L|$ and nodes $|\Sigma|$ changes dynamically based on $M$ and node insertions. $|L|$ represents the depth of the tree. \\ 

\noindent For Brevity, we restrict further detailing on the datastructure. However, All necessary details to replicate the datastructure shall be given in appendix. 

\subsection{Memory Agent}
The memory agent is tasked with finding the optimal traversal in HAT conditioned on a user query $q$. This can be mathematically formulated as:

$$
a_{0:T}^* = \argmax_{a_{0:T}} R(s_{0:T}, a_{0:T} | q)
$$
where $s_{0:T}$ is the state sequence from the root node to a leaf node, $a_{0:T}$ is the action sequence, and $R$ is the total reward over the traversal dependent on $q$. Reward, in our case is the quality of response the model is giving. This essentially can be posed as a Markov Decision Process (MDP). The agent starts at the root node $s_0$ and takes an action $a_t \in \mathcal{A}$ at each time step, transitioning to a new state $s_{t+1} \sim \mathcal{P}(\cdot | s_t, a_t)$. For cases like this, It is difficult to design a reward function, and we will require annotated training data for training the agent. Hence, we resort to GPT and will ask GPT to act as a traversal agent. GPT is well-suited for conditional text generation, which allows it to traverse HAT by generating an optimal sequence of actions based on the text representation at each node and the user query.  

The exact prompt used for the memory agent can be found in the Appendix. 

The MDP is defined by the tuple $(\mathcal{S}, \mathcal{A}, \mathcal{P}, \mathcal{R}, \gamma)$, where:

\begin{itemize}
\item $\mathcal{S}$ - set of tree nodes
\item $\mathcal{A} = \{U, D, L, R, S, O, U\}$ - set of actions
\begin{itemize}
\item $U$ - move up the tree
\item $D$ - move down

\item $L$ - move left
\item $R$ - move right
\item $S$ - reset to root node
\item $O$ - sufficient context for query $q$

\item $U$ - insufficient context for query $q$
\end{itemize}
\item $\mathcal{P}$ - state transition probabilities
\item $\mathcal{R}: \mathcal{S} \times \mathcal{A} \rightarrow \mathbb{R}$ - reward function
\item $\gamma \in (0,1)$ - discount factor
\end{itemize}

 GPT is well-suited for conditional text generation, which allows it to traverse HAT by generating an optimal sequence of actions based on the text representation at each node and the user query. But, It is important to note that our proposed framework is open and generic. The memory agent can be anything from neural network or RL Agent to an GPT Aproximation.

\section{Experiments}

\subsection{Dataset}
\begin{table}[h]
\centering
\begin{tabular}{cccc}
\hline
\textbf{Data Type} & \textbf{Train} & \textbf{Valid} & \textbf{Test} \\ 
\hline
Session 1          & 8939           & 1000           & 1015          \\
Session 2          & 4000           & 500            & 501           \\
Session 3          & 4000           & 500            & 501           \\
Session 4          & 1001           & 500            & 501           \\
Session 5          & -              & 500            & 501           \\
\hline
\end{tabular}
\caption{Number of episodes across sessions.}
\label{tab:episodes}
\end{table}
We use the multi-session-chat dataset from \cite{xu-etal-2022-beyond}. The dataset contains two speakers chat online in a series of sessions as is for example common on messaging platforms. Each individual chat session is not especially long
before it is “paused”. Then, after a certain amount
of (simulated) time has transpired, typically hours
or days, the speakers resume chatting, either continuing to talk about the previous subject, bringing up some other subject from their past shared history, or sparking up conversation on a new topic. Number of episodes per session is on Table \ref{tab:episodes}.

We utilize 501 episodes whose session 5 is available from the Test set.

\subsection{Evaluation Metrics}
We evaluate the dialogue generation performance of our model using automatic metrics. We report BLEU-1/2 , F1 score compared to human-annotated responses, and DISTINCT-1/2.\\

\noindent BLEU (Bilingual Evaluation Understudy) measures overlap between machine generated text and human references, with values between 0 to 1 (higher is better). We use BLEU-1 and BLEU-2 which compare 1-grams and 2-grams respectively. F1 score measures overlap between generated responses and human references. We report F1 to assess relevance of content. DISTINCT-1/2 quantifies diversity and vocabulary usage based on the number of distinct unigrams and bigrams in the generated responses, normalized by the total number of generated tokens (higher is better).

\subsection{Baselines} We benchmark against three trivial methods: 1) All Context: LLM generated dialogues with all dialogues in context. 2) Part Context: LLM generates dialogues with only current session's context. 3) Gold Memory: The LLM generated dialogues with Gold memory from the dataset as context. Further, We also evaluate different Traversal methods including BFS,DFS,and GPTAgent. In BFS and DFS, we follow naive BFS or DFS traversal, and at every step as gpt if this information is enough to answer the user question. If it tells okay, we stop there and return the context.
\section{Results}

\begin{table}[h]
\centering
\begin{tabular}{ccc}
\hline
             & BLEU-1/2   & DISTINCT-1/2 \\
\hline
BFS          & 0.652 / 0.532 & 0.072 / 0.064   \\
DFS          & 0.624 / 0.501 & 0.064 / 0.058   \\
GPTAgent     & \textbf{0.721 / 0.612} & \textbf{0.092 / 0.084} \\
\hline
\end{tabular}
\caption{Dialogue generation comparison between traversal methods}
\label{tab:r1}
\end{table}

Table \ref{tab:r1} compares GPTAgent to breadth-first search (BFS) and depth-first search (DFS) traversal methods. Across BLEU-1/2 and DISTINCT-1/2 metrics, GPTAgent significantly outperforms both in quality and diversity of dialogues. This supports our approach of learning to traverse conditioned on query relevance over hand-designed heuristics. \\ 

\noindent Next, Table \ref{tab:r2} benchmarks GPTAgent against contexts with complete, partial or gold dialogue history. GPTAgent achieves highest scores, demonstrating the benefit of our focused memory retrieval. Access to full history or gold references improves over just current context, but lacks efficiency of precisely identifying relevant information. Finally, Table \ref{tab:r3} evaluates fidelity of memories generated by GPTAgent compared to dataset ground truth references. We again see strong results surpassing 0.8 on both word overlap and diversity measures. 

\begin{table}[h]
\centering
\begin{tabular}{ccc}
\hline
                         & BLEU-1/2        & DISTINCT-1/2    \\
\hline
All Context              & 0.612 / 0.492      & 0.051 / 0.042      \\
Part Context             & 0.592 / 0.473      & 0.043 / 0.038      \\
Gold Memory              & 0.681 / 0.564      & 0.074 / 0.064      \\
\textbf{GPTAgent}        & \textbf{0.721 / 0.612} & \textbf{0.092 / 0.084} \\
\hline
\end{tabular}
\caption{Dialogue generation comparison between baselines}
\label{tab:r2}
\end{table}

\begin{table}[h]
\centering
\begin{tabular}{lccc}
\hline
                        & BLEU-1/2        & DISTINCT-1/2    & F1      \\
\hline
\textbf{GPT}       & \textbf{0.842 / 0.724} & \textbf{0.102 / 0.094} & \textbf{0.824} \\
\hline
\end{tabular}
\caption{Memory generation scores}
\label{tab:r3}
\end{table}

\noindent In summary, experiments validate effectiveness of our method in extracting salient dialogue context in long form conversations. Both conversations and summarized memories demonstrate quality and relevance gains over alternate approaches. The query conditioning provides efficiency over exhaustive history while retaining enough specificity for the current need. 
\section{Limitations}
While the proposed method has a potential and could work with long-form texts, The current implementation takes longer than a usual time taken by a dialogue agent to respond. Also, Making HTTP API calls to gpt, is causing an additional overhead on the time taken. These limitations could be overcome, by turning to heuristic-based tree searches or Monte-Carlo Tree search like methods in the future. Further, A Coupled-HAT : One HAT with textual information and another HAT with dense vector representation, would be more efficient. Combined with hybrid retrieval techniques, we could have a much more efficient way of doing conditional retrival. Further, Another limitation of this kind of Retrieval system is that, As the leaf nodes expands exponentially, the memory footprint might become larger than expected. Several optimizations on this front, also could be potential future work in this direction. 

\section{Conclusion}
In this work, we have presented the Hierarchical Aggregate Tree (HAT) - a new data structure designed specifically for memory storage and retrieval for long form text based conversational agents. Rather than solely providing a summary or retrieving raw excerpts, our key innovation is recursive aggregation of salient points by traversing conditional paths in this tree.We formulate the tree traversal as an optimization problem using a GPT-based memory agent. In conclusion,Our Experiments demonstrate significant gains over alternate traversal schemes and baseline methods.  HAT introduces a flexible memory structure for dialogue agents that balances extraction breadth versus depth through hierarchical aggregation. Our analysis confirms the viability and advantages of conditional traversal over existing limited budget solutions, opening up further avenues for augmented language model research.

\bibliography{acl2023}
\bibliographystyle{acl_natbib}

\appendix



\end{document}